\documentclass[11pt,a4paper]{article}
\usepackage[hyperref]{acl2020}
\usepackage{times}
\usepackage{latexsym}
\usepackage{amsfonts,amssymb}
\usepackage{amsmath}
\usepackage{booktabs}
\usepackage{graphicx}
\usepackage{subcaption}

\aclfinaltrue

\usepackage{microtype}

\title{\textbf{Every Document Owns Its Structure: Inductive Text Classification via Graph Neural Networks}}

\author{\textbf{Yufeng Zhang$^1$\footnotemark[1] , Xueli Yu$^1$\footnotemark[1] , Zeyu Cui$^1$, Shu Wu$^1$, Zhongzhen Wen$^2$ and Liang Wang$^1$} \\
  $^1$Institute of Automation, Chinese Academy of Sciences \\
  $^2$Xi'an Jiaotong University \\
  \texttt{\{yufeng.zhang,xueli.yu\}@cripac.ia.ac.cn} \\ \texttt{\{zeyu.cui,shu.wu\}@nlpr.ia.ac.cn} \\
  \texttt{burning21@stu.xjtu.edu.cn, wangliang@nlpr.ia.ac.cn} \\} 
  
\date{}

\begin{document}
\maketitle

\renewcommand{\thefootnote}{\fnsymbol{footnote}}
\footnotetext[1]{The first two authors contribute equally to this work.} 
\renewcommand{\thefootnote}{\arabic{footnote}}

\begin{abstract}
Text classification is fundamental in natural language processing (NLP), and Graph Neural Networks (GNN) are recently applied in this task. However, the existing graph-based works can neither capture the contextual word relationships within each document nor fulfil the inductive learning of new words. In this work, to overcome such problems, we propose TextING\footnote{https://github.com/CRIPAC-DIG/TextING} for inductive text classification via GNN. We first build individual graphs for each document and then use GNN to learn the fine-grained word representations based on their local structures, which can also effectively produce embeddings for unseen words in the new document. Finally, the word nodes are incorporated as the document embedding. Extensive experiments on four benchmark datasets show that our method outperforms state-of-the-art text classification methods. 
\end{abstract}

\section{Introduction}

Text classification is one of the primary tasks in the NLP field, as it provides fundamental methodologies for other NLP tasks, such as spam filtering, sentiment analysis, intent detection, and so forth.  Traditional methods for text classification include Naive Bayes \citep{androutsopoulos2000evaluation}, k-Nearest Neighbor \citep{tan2006effective} and Support Vector Machine \citep{forman2008bns}. They are, however, primarily dependent on the hand-crafted features at the cost of labour and efficiency.

There are several deep learning methods proposed to address the problem, among which Recurrent Neural Network (RNN) \citep{mikolov2010recurrent} and Convolutional Neural Network (CNN) \citep{kim2014convolutional} are essential ones. Based on them, extended models follow to leverage the classification performance, for instance, TextCNN \citep{kim2014convolutional}, TextRNN \citep{liu2016recurrent} and TextRCNN \citep{lai2015recurrent}. Yet they all focus on the locality of words and thus lack of long-distance and non-consecutive word interactions. Graph-based methods are recently applied to solve such issue, which do not treat the text as a sequence but as a set of co-occurrent words instead. For example, \citet{yao2019graph} employ Graph Convolutional Networks \citep{kipf2017semi} and turns the text classification problem into a node classification one (TextGCN). Moreover, \citet{huang2019text} improve TextGCN by introducing the message passing mechanism and reducing the memory consumption.

However, there are two major drawbacks in these graph-based methods. First, the contextual-aware word relations within each document are neglected. To be specific, TextGCN \citep{yao2019graph} constructs a single graph with global relations between documents and words, where fine-grained text level word interactions are not considered \citep{wu2019session,hu2019graphair,hu2019hierarchical}. In \citet{huang2019text}, the edges of the graph are globally fixed between each pair of words, but the fact is that they may affect each other differently in a different text. Second, due to the global structure, the test documents are mandatory in training. Thus they are inherently \emph{transductive} and have difficulty with \emph{inductive} learning, in which one can easily obtain word embeddings for new documents with new structures and words using the trained model.

Therefore, in this work, we propose a novel Text classification method for INductive word representations via Graph neural networks, termed TextING. In contrast to previous graph-based approaches with global structure, we train a GNN that can depict the detailed word-word relations using only training documents, and generalise to new documents in test. We build individual graphs by applying the sliding window inside each document \citep{rousseau2015text}. The information of word nodes is propagated to their neighbours via the Gated Graph Neural Networks \citep{li2015gated,li2019fi}, which is then aggregated into the document embedding. We also conduct extensive experiments to examine the advantages of our approach against baselines, even when words in test are mostly unseen (21.06\% average gain in such inductive condition).  Noticing a concurrent work \citep{nikolentzos2019message} also reinforces the approach with a similar graph network structure, we describe the similarities and differences in the method section. To sum up, our contributions are threefold:
\begin{itemize}
    \item We propose a new graph neural network for text classification, where each document is an individual graph and text level word interactions can be learned in it.
    \item Our approach can generalise to new words that absent in training, and it is therefore applicable for inductive circumstances.
    \item We demonstrate that our approach outperforms state-of-the-art text classification methods experimentally.
\end{itemize}

\section{Method}

\begin{figure}[ht]
\setlength{\belowcaptionskip}{-0.5cm}
\setlength{\abovecaptionskip}{0.2cm}
\center{\includegraphics[width=7cm]  {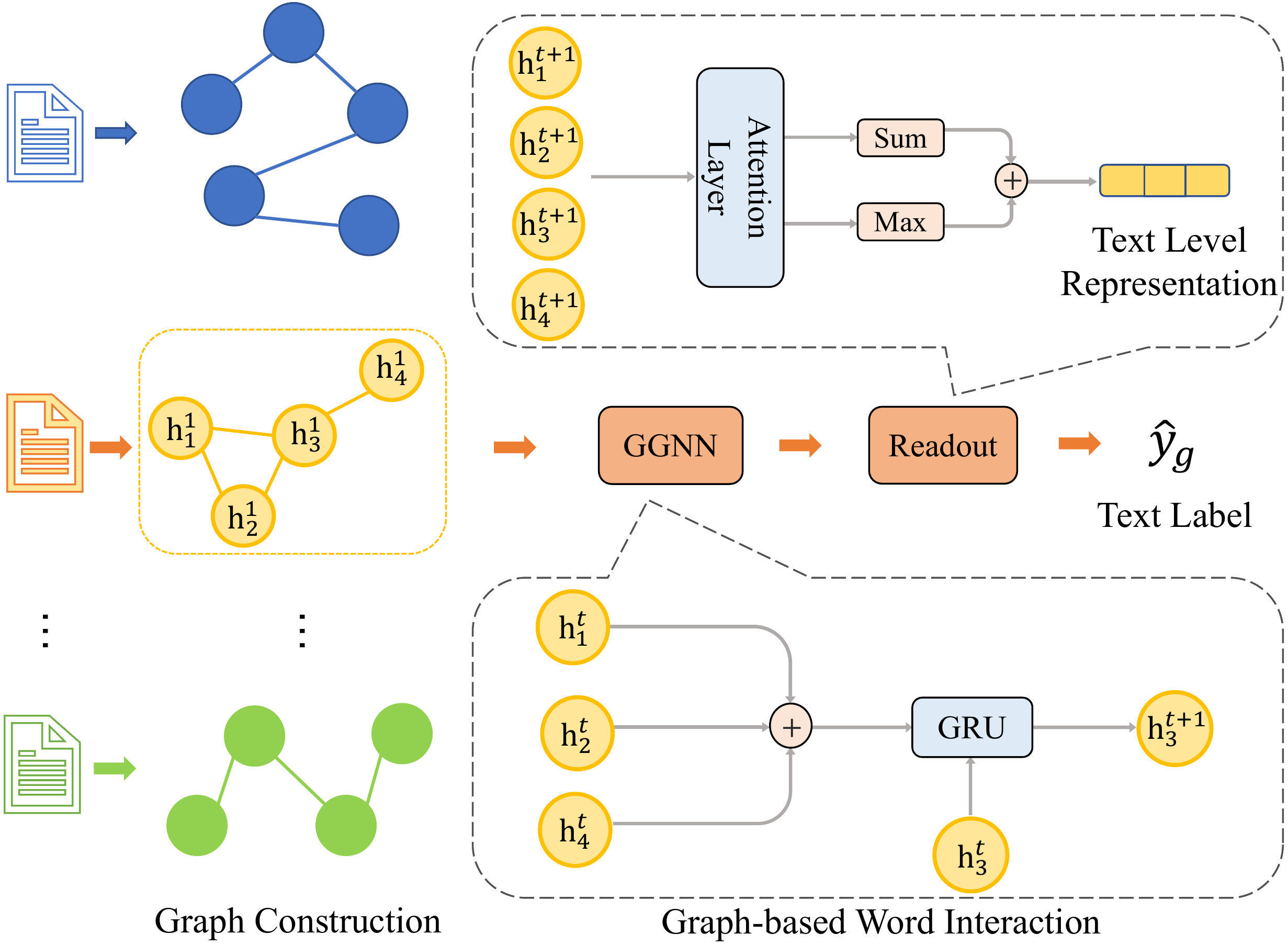}} \caption{\label{architecture} The architecture of TextING. As an example, upon a graph of document, every word node updates itself from its neighbours and they aggregate to the ultimate graph representation.}
\end{figure}

TextING comprises three key components: the graph construction, the graph-based word interaction, and the readout function. The architecture is illustrated in Figure~\ref{architecture}. In this section, we detail how to implement the three and how they work.

\subsection*{Graph Construction}
We construct the graph for a textual document by representing unique words as vertices and co-occurrences between words as edges, denoted as $\mathcal{G}=(\mathcal{V}, \mathcal{E})$ where $\mathcal{V}$ is the set of vertices and $\mathcal{E}$ the edges. The co-occurrences describe the relationship of words that occur within a fixed-size sliding window (length 3 at default) and they are undirected in the graph. \citet{nikolentzos2019message} also use a sliding window of size 2. However, they include a particular master node connecting to every other node, which means the graph is densely connected and the structure information is vague during message passing. 

The text is preprocessed in a standard way, including tokenisation and stopword removal \citep{blanco2012graph,rousseau2015text}. Embeddings of the vertices are initialised with word features, denoted as {\textbf h} $\in \mathbb{R}^{|\mathcal{V}| \times d}$ where $d$ is the embedding dimension. Since we build individual graphs for each document, the word feature information is propagated and incorporated contextually during the word interaction phase.

\subsection*{Graph-based Word Interaction}
Upon each graph, we then employ the Gated Graph Neural Networks \citep{li2015gated} to learn the embeddings of the word nodes. A node could receive the information {\textbf a} from its adjacent neighbours and then merge with its own representation to update. As the graph layer operates on the first-order neighbours, we can stack such layer $t$ times to achieve high-order feature interactions, where a node can reach another node $t$ hops away. The formulas of the interaction are:
\begin{align}
    {\rm \textbf a}^t &= {\rm \textbf A}{\rm \textbf h}^{t-1}{\rm \textbf W}_a,~ \\
    {\rm \textbf z}^t &= \sigma\left({\rm \textbf W}_z{\rm \textbf a}^t + {\rm \textbf U}_z{\rm \textbf h}^{t-1}+{\rm \textbf b}_z \right),~ \\
    {\rm \textbf r}^t &= \sigma\left({\rm \textbf W}_r{\rm \textbf a}^t + {\rm \textbf U}_r{\rm \textbf h}^{t-1}+{\rm \textbf b}_r \right),~ \\
    {\rm \tilde{\textbf h}}^t &= {\rm tanh}\left({\rm \textbf W}_h{\rm \textbf a}^t + {\rm \textbf U}_h({\rm \textbf r}^t \odot {\rm \textbf h}^{t-1})+{\rm \textbf b}_h\right),~ \\
    {\rm \textbf h}^t &= {\rm \tilde{\textbf h}}^t \odot {\rm \textbf z}^t + {\rm \textbf h}^{t-1} \odot \left(1-{\rm \textbf z}^t\right),~
\end{align}
where {\textbf A} $\in \mathbb{R}^{|\mathcal{V}| \times |\mathcal{V}|}$ is the adjacency matrix, $\sigma$ is the sigmoid function, and all {\textbf W}, {\textbf U} and {\textbf b} are trainable weights and biases. \textbf{z} and \textbf{r} function as the update gate and reset gate respectively to determine to what degree the neighbour information contributes to the current node embedding.

\begin{table*}
\setlength{\abovecaptionskip}{0.1cm}
\centering
\footnotesize
\caption{\label{statistic} The statistics of the datasets including both short (sentence) and long (paragraph) documents. The vocab means the number of unique words in a document. The Prop.NW denotes the proportion of new words in test.}
\begin{tabular}{lcccrrrrr}
\toprule \textbf{Dataset} & \textbf{\# Docs} & \textbf{\# Training} & \textbf{\# Test} & \textbf{\# Classes} & \textbf{Max.Vocab} & \textbf{Min.Vocab} & \textbf{Avg.Vocab} & \textbf{Prop.NW}\\ \midrule
MR & 10,662 & 7,108 & 3,554 & 2 & 46 & 1 & 18.46 & 30.07\%\\
R8 & 7,674 & 5,485 & 2,189 & 8 & 291 & 4 & 41.25 & 2.60\%\\
R52 & 9,100 & 6,532 & 2,568 & 52 & 301 & 4 & 44.02 & 2.64\%\\
Ohsumed & 7,400 & 3,357 & 4,043 & 23 & 197 & 11 & 79.49 & 8.46\%\\
\bottomrule
\end{tabular}
\end{table*}

\subsection*{Readout Function}
After the word nodes are sufficiently updated, they are aggregated to a graph-level representation for the document, based on which the final prediction is produced. We define the readout function as:
\begin{align}
    {\rm \textbf h}_v &= \sigma\left(f_1({\rm \textbf h}_v^t)\right) \odot {\rm tanh}\left(f_2({\rm \textbf h}_v^t)\right),~ \\
    {\rm \textbf h}_\mathcal{G} &= \frac{1}{|\mathcal{V}|}\sum_{v \in \mathcal{V}}{\rm \textbf h}_v + {\rm Maxpooling}\left({\rm \textbf h}_1 ...{\rm \textbf h}_\mathcal{V}\right), 
\end{align}
where $f_1$ and $f_2$ are two multilayer perceptrons (MLP). The former performs as a soft attention weight while the latter as a non-linear feature transformation. In addition to averaging the weighted word features, we also apply a max-pooling function for the graph representation {\textbf h}$_\mathcal{G}$. The idea behind is that every word plays a role in the text and the keywords should contribute more explicitly.

Finally, the label is predicted by feeding the graph-level vector into a softmax layer. We minimise the loss through the cross-entropy function:
\begin{align}
    \hat{y}_\mathcal{G} &= {\rm softmax}\left({\rm \textbf W \textbf h}_\mathcal{G} + {\rm\textbf b}\right),~ \\
    \mathcal{L} &= - \sum_i y_{\mathcal{G}i} {\rm log}\left(\hat{y}_{\mathcal{G} i}\right),~
\end{align}
where \textbf W and \textbf b are  weights and bias, and $y_{\mathcal{G}i}$ is the $i$-th element of the one-hot label.

\subsection*{Model Variant}
We also extend our model with a multichannel branch TextING-M, where graphs with local structure (original TextING) and graphs with global structure (subgraphs from TextGCN) work in parallel. The nodes remain the same whereas the edges of latter are extracted from the large graph (built on the whole corpus) for each document. We train them separately and make them vote 1:1 for the final prediction. Although it is not the inductive case, our point is to investigate whether and how the two could complement each other from micro and macro perspectives.

\begin{table*}
\setlength{\abovecaptionskip}{0.1cm}
\centering
\footnotesize
\caption{\label{results} Test accuracy (\%) of various models on four datasets. The mean $\pm$ standard deviation of our model is reported according to 10 times run. Note that some baseline results are from \citep{yao2019graph}.}
\begin{tabular}{lcccc}
\toprule \textbf{Model} & \textbf{MR} & \textbf{R8} & \textbf{R52} & \textbf{Ohsumed} \\ \midrule
CNN (Non-static) & 77.75 $\pm$ 0.72 & 95.71 $\pm$ 0.52 & 87.59 $\pm$ 0.48 & 58.44 $\pm$ 1.06 \\
RNN (Bi-LSTM) & 77.68 $\pm$ 0.86 & 96.31 $\pm$ 0.33 & 90.54 $\pm$ 0.91 & 49.27 $\pm$ 1.07 \\
fastText & 75.14 $\pm$ 0.20 & 96.13 $\pm$ 0.21 & 92.81 $\pm$ 0.09 & 57.70 $\pm$ 0.49 \\
SWEM & 76.65 $\pm$ 0.63 & 95.32 $\pm$ 0.26 & 92.94 $\pm$ 0.24 & 63.12 $\pm$ 0.55 \\
TextGCN & 76.74 $\pm$ 0.20 & 97.07 $\pm$ 0.10 & 93.56 $\pm$ 0.18 & 68.36 $\pm$ 0.56 \\
\citet{huang2019text} & - & 97.80 $\pm$ 0.20 & 94.60 $\pm$ 0.30 & 69.40 $\pm$ 0.60 \\ \midrule
TextING & 79.82 $\pm$ 0.20 & 98.04 $\pm$ 0.25 & 95.48 $\pm$ 0.19 & 70.42 $\pm$ 0.39 \\
TextING-M & 80.19 $\pm$ 0.31 & 98.13 $\pm$ 0.12 & 95.68 $\pm$ 0.35 & 70.84 $\pm$ 0.52 \\
\bottomrule
\end{tabular}
\end{table*}

\section{Experiments}
In this section, we aim at testing and evaluating the overall performance of TextING. During the experimental tests, we principally concentrate on three concerns: (\romannumeral1) the performance and advantages of our approach against other comparable models, (\romannumeral2) the adaptability of our approach for words that are never seen in training, and (\romannumeral3) the interpretability of our approach on how words impact a document.

\paragraph{Datasets.}
For the sake of consistency, we adopt four benchmark tasks the same as in \citep{yao2019graph}: (\romannumeral1) classifying movie reviews into positive or negative sentiment polarities (MR)\footnote{http://www.cs.cornell.edu/people/pabo/movie-review-data/}, (\romannumeral2) \& (\romannumeral3) classifying documents that appear on Reuters newswire into 8 and 52 categories (R8 and R52 respectively)\footnote{http://disi.unitn.it/moschitti/corpora.htm}, (\romannumeral4) classifying medical abstracts into 23 cardiovascular diseases categories (Ohsumed)\footnote{https://www.cs.umb.edu/\~{}smimarog/textmining/datasets/}. Table~\ref{statistic} demonstrates the statistics of the datasets as well as their supplemental information. 

\paragraph{Baselines.} 
We consider three types of models as baselines: (\romannumeral1) traditional deep learning methods including TextCNN \citep{kim2014convolutional} and TextRNN \citep{liu2016recurrent}, (\romannumeral2) simple but efficient strategies upon word features including fastText \citep{joulin2017bag} and SWEM \citep{shen2018baseline}, and (\romannumeral3) graph-based methods for text classification including TextGCN \citep{yao2019graph} and \citet{huang2019text}.

\paragraph{Experimental Set-up.}
For all the datasets, the training set and the test set are given, and we randomly split the training set into the ratio 9:1 for actual training and validation respectively. The hyperparameters were tuned according to the performance on the validation set. Empirically, we set the learning rate as 0.01 with Adam \citep{kingma2015adam} optimiser and the dropout rate as 0.5. Some depended on the intrinsic attributes of the dataset, for example, the word interaction step and the sliding window size. We refer to them in the parameter sensitivity subsection.

Regarding the word embeddings, we used the pre-trained GloVe \citep{pennington2014glove}\footnote{http://nlp.stanford.edu/data/glove.6B.zip} with $d=300$ as the input features while the out-of-vocabulary (OOV) words' were randomly sampled from a uniform distribution [-0.01, 0.01]. For a fair comparison, the other baseline models shared the same embeddings. 

\paragraph{Results.} Table~\ref{results} presents the performance of our model as well as the baselines. We observe that graph-based methods generally outperform other types of models, suggesting that the graph model benefits to the text processing. Further, TextING ranks top on all tasks, suggesting that the individual graph exceeds the global one. Particularly, the result of TextING on MR is remarkably higher. Because the short documents in MR lead to a low-density graph in TextGCN, it restrains the label message passing among document nodes, whereas our individual graphs (documents) do not rely on such label message passing mechanism. Another reason is that there are approximately one third new words in test as shown in Table~\ref{statistic}, which implies TextING is more friendly to unseen words. The improvement on R8 is relatively subtle since R8 is simple to fit and the baselines are rather satisfying. The proportion of new words is also low on R8.

The multichannel variant also performs well on all datasets. It implies the model can learn different patterns through different channels.

\paragraph{Under Inductive Condition.} 
To examine the adaptability of TextING under inductive condition, we reduce the amount of training data to 20 labelled documents per class and compare it with TextGCN. Word nodes absent in the training set are masked for TextGCN to simulate the inductive condition. In this scenario, most of the words in the test set are unseen during training, which behaves like a rigorous cold-start problem. The result of both models on MR and Ohsumed are listed in Table~\ref{inductive}. An average gain of 21.06\% shows that TextING is much less impacted by the reduction of exposed words. In addition, a tendency of test performance and gain with different percentages of training data on MR is illustrated as Figure~\ref{fig:inductive_res}. TextING shows a consistent improvement when increasing number of words become unseen.

\begin{table}
\setlength{\abovecaptionskip}{0.1cm}
\centering
\footnotesize
\caption{\label{inductive} Accuracy (\%) of TextGCN and TextING on MR and Ohsumed, where MR uses 40 labelled documents (0.5\% of full training data) and Ohsumed uses 460 labelled documents (13.7\% of full training data).}
\begin{tabular}{lcc}
\toprule \textbf{Model} & \textbf{MR*} & \textbf{Ohsumed*} \\ \midrule
TextGCN & 53.15 & 47.24 \\
TextING & 64.43 & 57.11 \\ \midrule
\# Words in Training & 465 & 7,009 \\
\# New Words in Test & 18,299 & 7,148 \\
\bottomrule
\end{tabular}
\end{table}

\begin{figure}[ht]
\setlength{\abovecaptionskip}{0.1cm}
\setlength{\belowcaptionskip}{-0.3cm}
    \centering
    \includegraphics[scale=0.45]{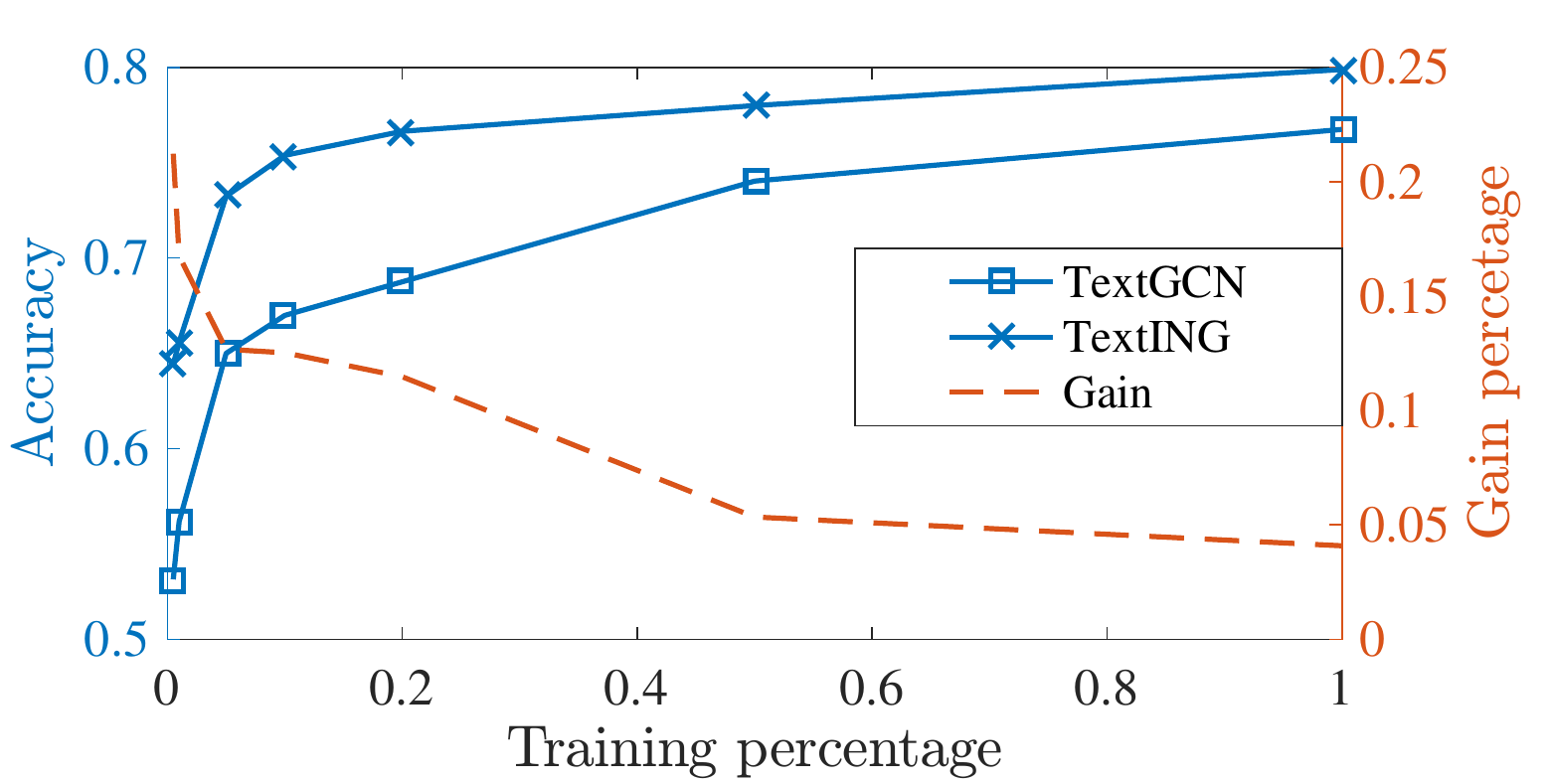}

    \caption{Test performance and gain with different percent of training data ranging from 0.005 to 1 on MR. The less data in training, the more new words in test.}
    \label{fig:inductive_res}
\end{figure}

\begin{figure}[ht]
\setlength{\belowcaptionskip}{-0.3cm}
\begin{subfigure}{0.22\textwidth}
  \centering
  \includegraphics[width=1.0\linewidth]{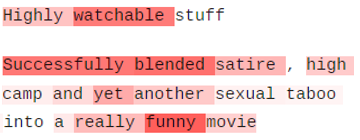}  
  \caption{Positive reviews}
  \label{fig:sub-first}
\end{subfigure}
\begin{subfigure}{0.22\textwidth}
  \centering
  \includegraphics[width=1.0\linewidth]{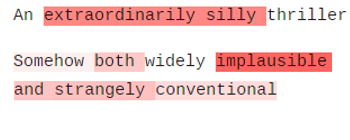}  
  \caption{Negative reviews}
  \label{fig:sub-second}
\end{subfigure}
\caption{Attention visualisation of positive and negative movie reviews in MR.}
\label{fig:attention}
\end{figure}

\paragraph{Case Study.} 
To understand what is of importance that TextING learns for a document, we further visualise the attention layer (i.e. the readout function), illustrated as Figure~\ref{fig:attention}. The highlighted words are proportional to the attention weights, and they show a positive correlation to the label, which interprets how TextING works in sentiment analysis.

\begin{figure}[ht]
\setlength{\belowcaptionskip}{-0.2cm}
\begin{subfigure}{.23\textwidth}
  \centering
  \includegraphics[width=1\linewidth]{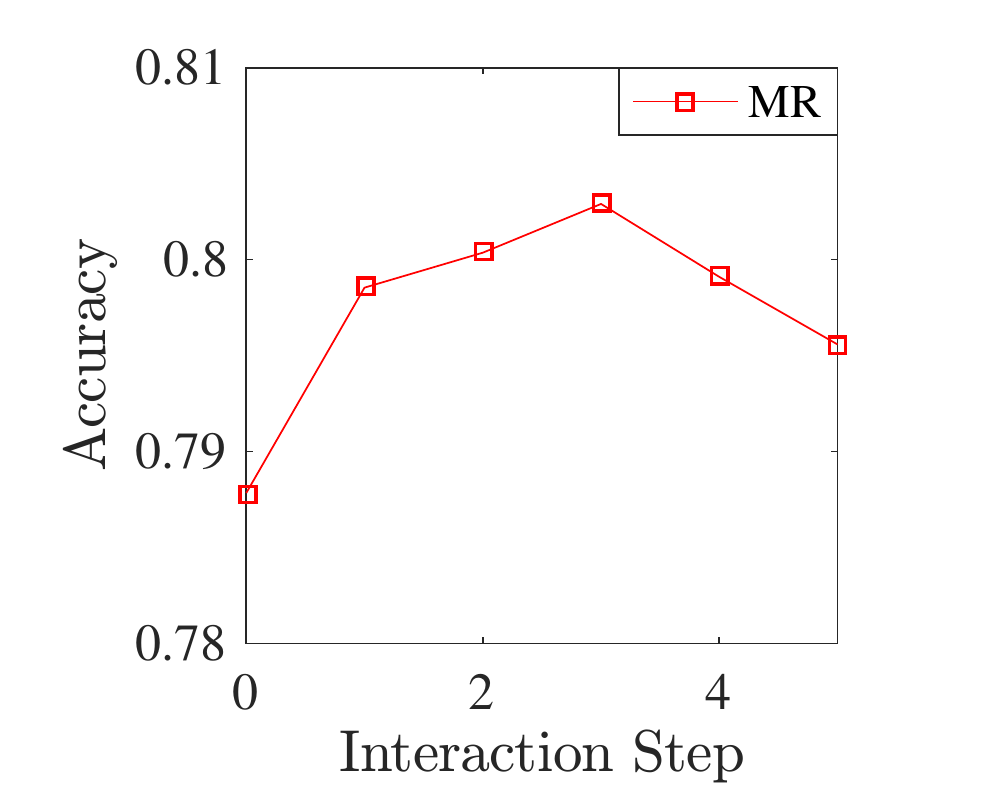}  
  \caption{MR}
  \label{fig:sub-first}
\end{subfigure}
\begin{subfigure}{.23\textwidth}
  \centering
  \includegraphics[width=1\linewidth]{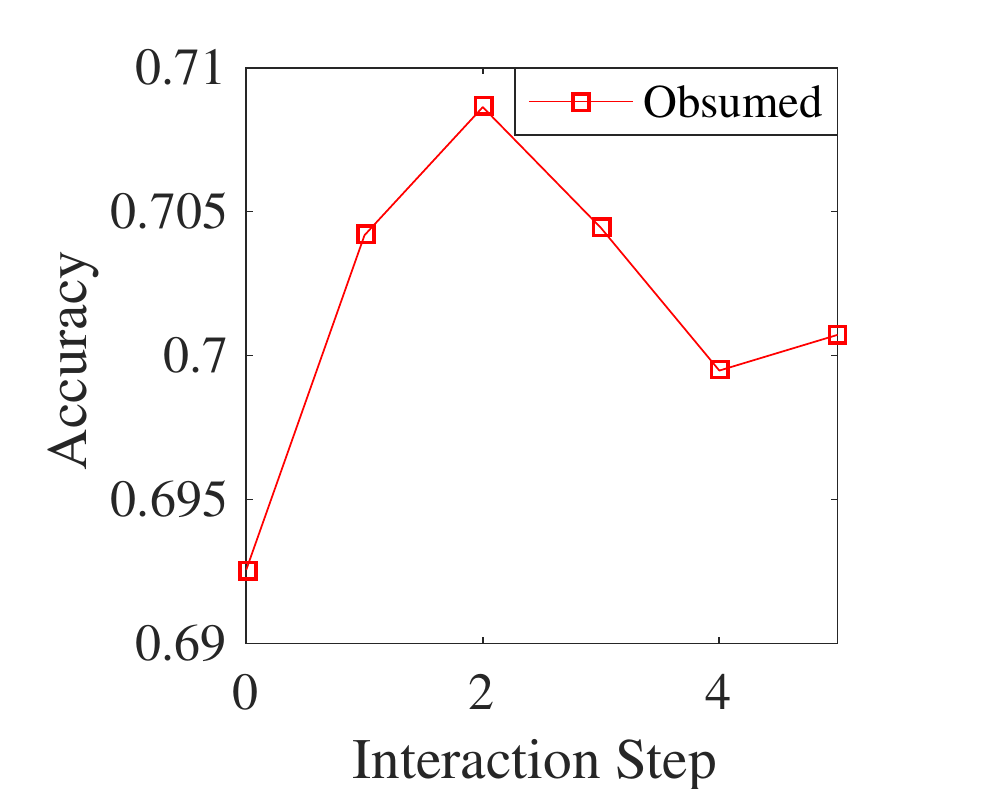}  
  \caption{Ohsumed}
  \label{fig:sub-second}
\end{subfigure}

\caption{Accuracy with varying interaction steps.}
\label{fig:interaction_step}
\end{figure}

\begin{figure}[ht]
\setlength{\belowcaptionskip}{-0.2cm}
\begin{subfigure}{.23\textwidth}
  \centering
  \includegraphics[width=1\linewidth]{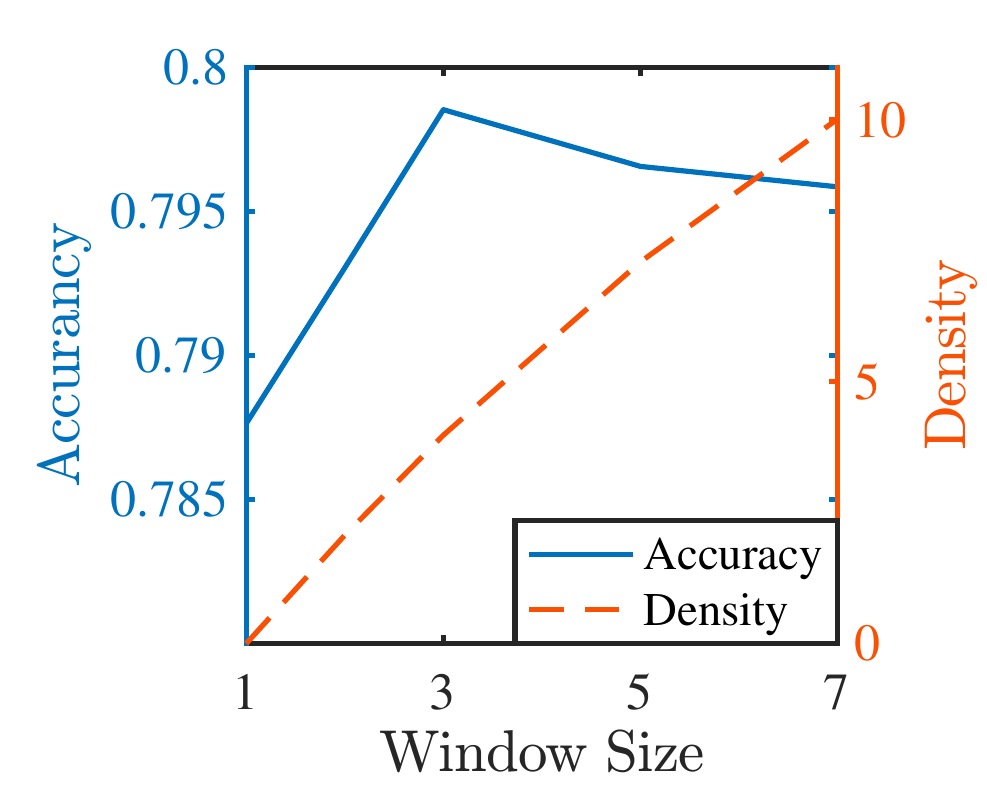}  
  \caption{MR}
  \label{fig:sub-first}
\end{subfigure}
\begin{subfigure}{.23\textwidth}
  \centering
  \includegraphics[width=1\linewidth]{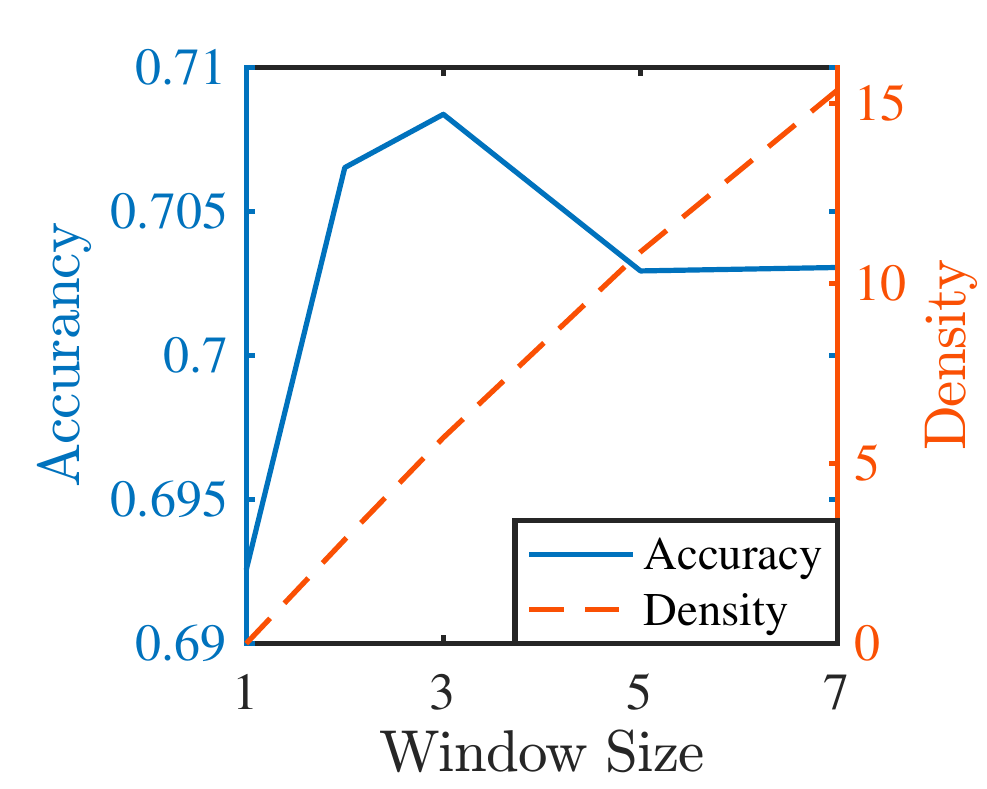}  
  \caption{Ohsumed}
  \label{fig:sub-second}
\end{subfigure}

\caption{Accuracy with varying graph density.}
\label{fig:window_size}
\end{figure}

\paragraph{Parameter Sensitivity.}
 Figure~\ref{fig:interaction_step} exhibits the performance of TextING with a varying number of the graph layer on MR and Ohsumed. The result reveals that with the increment of the layer, a node could receive more information from high-order neighbours and learn its representation more accurately. Nevertheless, the situation reverses with a continuous increment, where a node receives from every node in the graph and becomes over-smooth. 
 Figure~\ref{fig:window_size} illustrates the performance as well as the graph density of TextING with a varying window size on MR and Ohsumed. It presents a similar trend as the interaction step's when the number of neighbours of a node grows.
 
\section{Conclusion}
We proposed a novel graph-based method for inductive text classification, where each text owns its structural graph and text level word interactions can be learned. Experiments proved the effectiveness of our approach in modelling local word-word relations and word significances in the text.

\section*{Acknowledgement}
This work is supported by National Natural Science Foundation of China (U19B2038, 61772528) and National Key Research and Development Program (2018YFB1402600, 2016YFB1001000).


\bibliography{anthology,acl2020}
\bibliographystyle{acl_natbib}

\end{document}